\def\year{2022}\relax
\documentclass[letterpaper]{article} 
\usepackage{aaai22}  
\usepackage{times}  
\usepackage{helvet}  
\usepackage{courier}  
\usepackage[hyphens]{url}  
\usepackage{graphicx} 
\usepackage{natbib}  
\usepackage{caption} 
\DeclareCaptionStyle{ruled}{labelfont=normalfont,labelsep=colon,strut=off} 
\usepackage{algorithm}
\usepackage{algorithmic}

\usepackage{newfloat}
\usepackage{listings}
\floatstyle{ruled}
\newfloat{listing}{tb}{lst}{}
\floatname{listing}{Listing}
\title{Better Query Graph Selection for Knowledge Base Question Answering}

\author{
	Yonghui Jia, Wenliang Chen \\
	Institute of Artificial Intelligence, School of Computer Science and Technology, \\ Soochow University, China\\
	{\tt yhjia2018@163.com, wlchen@suda.edu.cn}
}

\begin{document}
\maketitle
\begin{abstract}
This paper presents a novel approach based on semantic parsing to improve the performance of Knowledge Base Question Answering (KBQA). Specifically, we focus on how to select an optimal query graph from a candidate set so as to retrieve the answer from knowledge base (KB). In our approach, we first propose to linearize the query graph into a sequence, which is used to form a sequence pair with the question. It allows us to use mature sequence modeling, such as BERT, to encode the sequence pair. Then we use a ranking method to sort candidate query graphs. In contrast to the previous studies, our approach can efficiently model semantic interactions between the graph and the question as well as rank the candidate graphs from a global view. The experimental results show that our system achieves the top performance on ComplexQuestions and the second best performance on WebQuestions.
\end{abstract}
\section{Introduction}
Knowledge Base Question Answering (KBQA) is a popular task defined to take natural language questions as input and return corresponding entities or attributes from knowledge bases, such as DBPedia~\citep{auer2007dbpedia} and Freebase~\citep{bollacker2008freebase}. One representative line of approaches to KBQA builds on semantic parsing (SP) that converts input questions into formal meaning representations and then transforms them into query languages like SPARQL~\citep{berant2013semantic,yih2015semantic,sun2020sparqa}. There are two types of SP-based solutions. One is using generic meaning representations, such as $\lambda-DCS$~\citep{liang2013lambda}. However, this type of solutions tends to suffer from the gap between the set of ontology or relationship in the meaning representations and the set in knowledge bases~\citep{kwiatkowski2013scaling}.

\begin{figure}
    \centering
    \includegraphics[width=7.5cm]{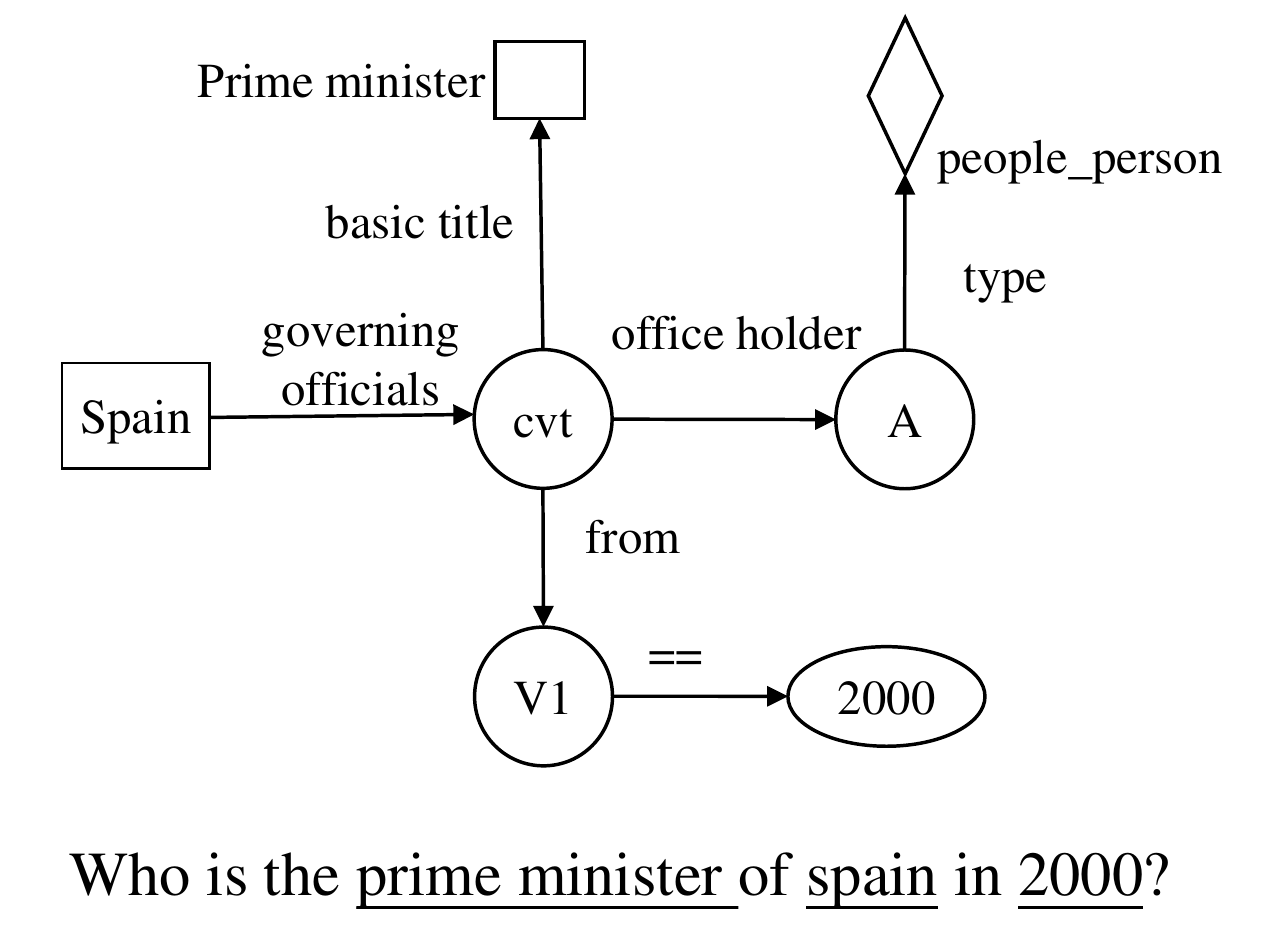}
    \caption{The example question and its corresponding query graph structure.}
    \label{fig:querygraph}
\end{figure}

Another type of solutions is to use query graph to represent the semantics of questions, which is supposed to be able to  overcome the issue mentioned above~\citep{yih2015semantic,bao2016constraint,hu2017answering}.
Figure~\ref{fig:querygraph} presents an illustrating query graph whose nodes and edges correspond to the entities and relationships in a knowledge base. By using query graph as the representation, the process of KBQA can be divided into two steps: query graph generation and query graph selection. The former step constructs a set of query graph candidates from the input question, while the latter decides the optimal query graph that is used to retrieve the final answer. We can see that the component of query graph selection is critical to the overall performance of KBQA systems.
\begin{figure*}
\centering
\includegraphics[width=17cm]{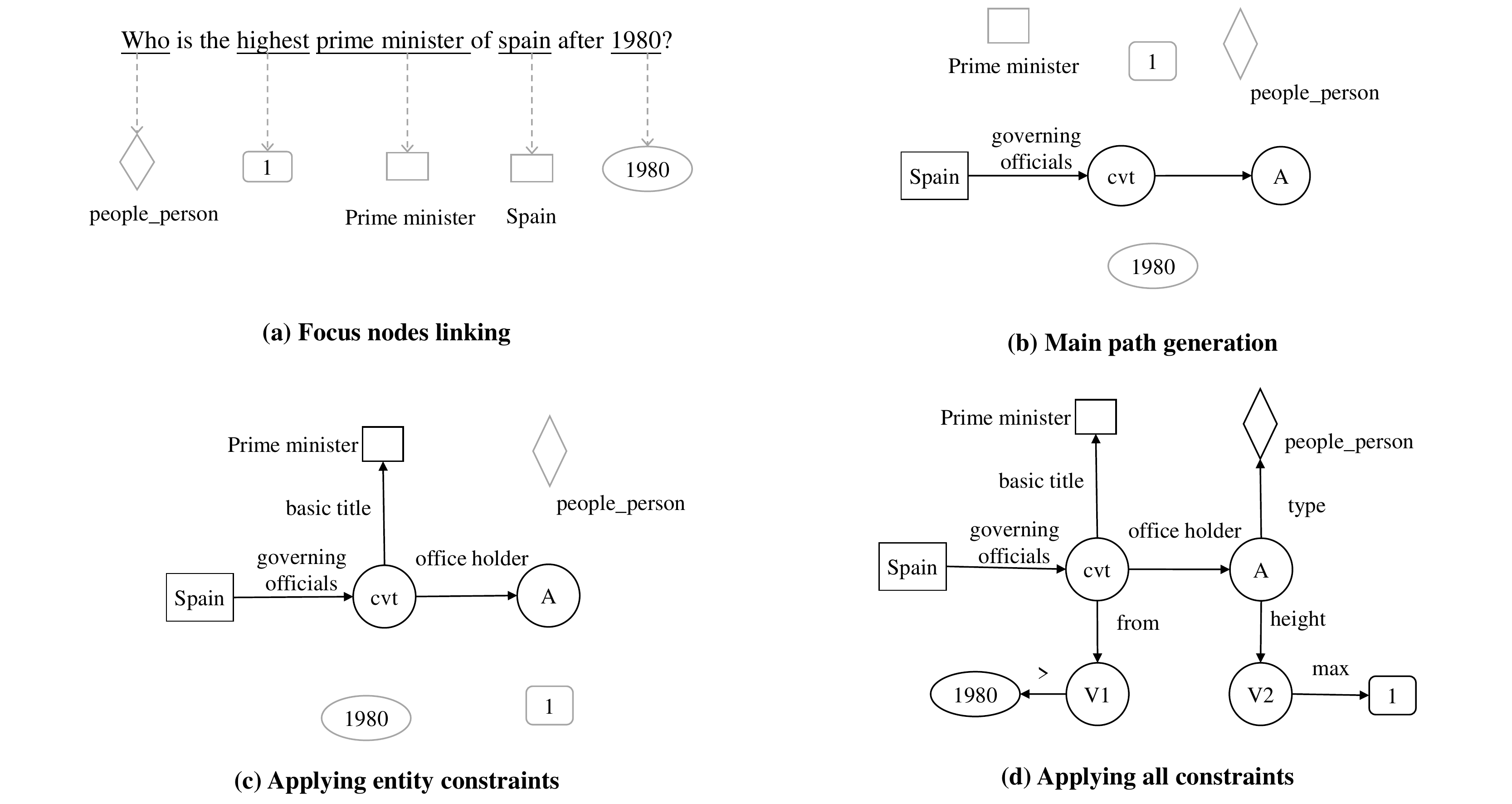}
\caption{Query graph generation process of ``Who is the highest prime minister of spain after 1980?''. Figure (a) shows the corresponding focus nodes linking result, Figure (b) shows the corresponding main path, Figure (c) shows the result after adding entity constraints to the main path, and Figure (d) shows the query graph after adding all constraints.}
\label{fig:graph_generation}
\end{figure*}

Query graph selection is essentially a matching task between the question and candidate query graphs. Existing systems put focus on encoding the query graphs with hand-crafted features~\citep{yih2015semantic,luo2018knowledge}. In these works, they first calculate the semantic similarity between the query graph and the question using \textit{cosine} similarity function. Then the similarity score is used as a feature, together with other local features to represent the query graph and the question. Finally, they feed the feature representation into a one-layer neural network model to obtain the score. Previous approaches have achieved a certain success for this task. However, we argue that 1) simply using \textit{cosine} distance function to measure the semantic similarity leads to the loss of interaction information between the graph and the question, and 2) the hand-crafted features are usually not robust and are often not necessary for deep neural networks. 

To address the above problems, in this paper we propose to translate the matching between the question and the query graph into the matching between two sequences for naturally modeling the interaction between the question and query graph. To this end, we linearize the query graphs into sequences. This strategy makes the question and linearized query graphs are both in sequence format, which is more convenient for using mature sequence modeling methods such as BERT~\citep{devlin2018bert} and GPT-3~\citep{brown2020language}. In addition, we select the optimal query graph with a ranking strategy, hoping to take the relationship between candidate query graphs into consideration. 
Inspired by learning to rank methods~\citep{li2011learning,pirtoacua2019answering,han2020learning}, we utilize the listwise strategy to sort candidate query graphs from a global view, instead of the pairwise strategy which is used in the previous study~\citep{luo2018knowledge}. Experimental results on two widely-used KBQA datasets demonstrate the effectiveness of our proposed approach: the best performance on ComplexQuestions and the second best on WebQuestions.

Overall, we make the following contributions.
\begin{itemize}
    \item We propose a novel approach for better query graph selection in KBQA. In our approach, we convert the query graph into the corresponding sequence format and thus the problem of matching between the question and the query graph is translated into the matching between two sequences. This allows us to use BERT to efficiently model interactions between the graph and the question. Moreover, our approach does not require any hand-crafted features.
    \item  In addition, we use the listwise strategy to sort the candidate query graphs which takes the relationship among the candidate graphs into consideration from a global view. Compared with the Pairwise Ranking used in \citep{luo2018knowledge}, Listwise Ranking achieves better performance on both two datasets.
\end{itemize}
\begin{figure*}
    \centering
    \includegraphics[width=17cm]{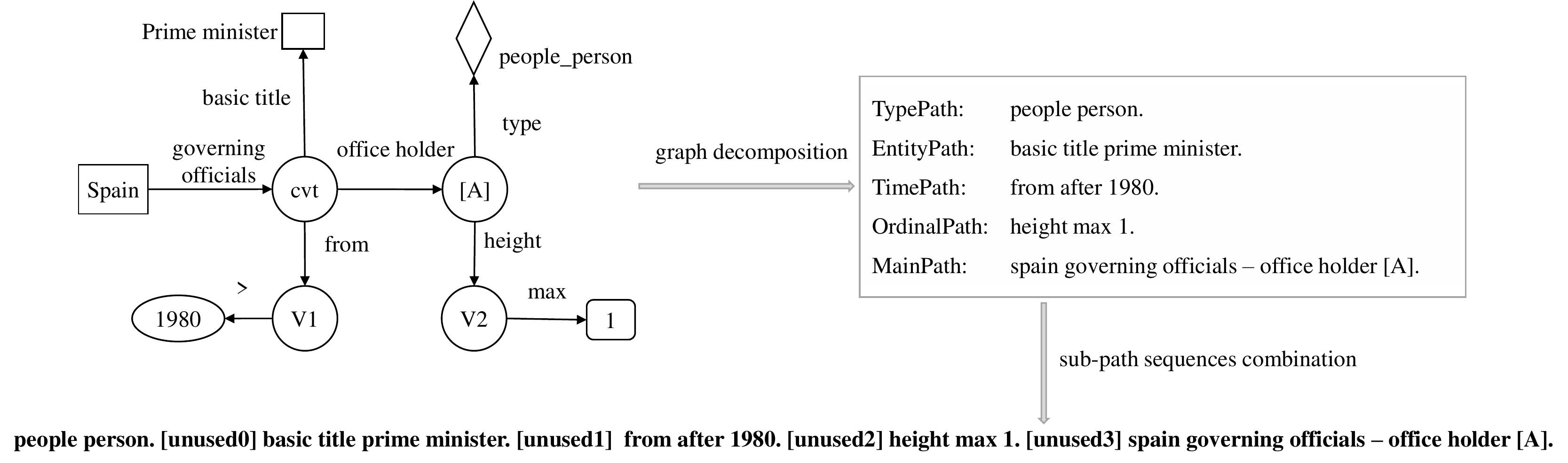}
    \caption{The process of converting query graph to sequence.}
    \label{fig:graph2sequence}
\end{figure*}

\section{Our Approach}
In this section, we describe our approach in detail. 
We divide the KBQA process into two subtasks: query graph generation and query graph selection. 
Formally, given a question $q$ and a knowledge base (KB), the semantics of $q$ is analyzed through the query graph generation, and a set of candidate query graphs $G=\{g_1, g_2, ..., g_n\}$ is obtained. Then, an optimal query graph $g^*$ is selected from the candidate set $G$ through the query graph selection. Finally, we convert $g^*$ into the SPARQL format to retrieve the final answer to question $q$.

Compared with the previous studies of using query graphs~\citep{yih2015semantic,hu2017answering,luo2018knowledge}, we improve our system by using a different solution on query graph selection.
The basic idea is that, we linearize each candidate query graph into a sequence and thus the problem of matching between the question and the candidate query graph becomes the matching between two sequences. To select the optimal query graph $g^*$, we propose a ranking method with the listwise strategy, hoping to take the relationship between the candidate query graphs into consideration from a global view.

\subsection{Query Graph Generation}\label{sec:qgg}


The goal of the query graph generation is to map the question into a semantic representation in the form of graphs.  In this step, we follow the procedure of the previous studies~\citep{yih2015semantic,luo2018knowledge} to generate the candidate query graphs.

Given question $q$, we first conduct focus nodes linking to identify four types of constraints in the question, which are \textbf{entity}, implicit \textbf{type}, \textbf{time} interval, and \textbf{ordinal}. For entity linking, we utilize the tool SMART~\citep{yang2016s} to obtain (mention, entity) pairs. For type linking, we use word embeddings to calculate the similarity between consecutive sub-sequences in the question (up to three words) and all the type words in the knowledge base, and select the top-10 (mention, type) pairs according to the similarity scores. Regarding time word linking, we use regular expression matching to extract time information. As for ordinal number linking, we use a predefined ordinal vocabulary and ``ordinal number+superlative'' pattern to extract the integer expressions.~\footnote{About 20 superlative words, such as largest, highest, latest.} We also use the entity enrichment method in ~\citet{luo2018knowledge} to improve focus nodes linking. Figure~\ref{fig:graph_generation}(a) shows an example after focus nodes linking. 

After focus nodes linking, we get the main path by performing a one-hop and two-hop search based on the linked entity words, as shown in Figure~\ref{fig:graph_generation}(b). Next, entity constraints are added to the nodes in the main path. Figure~\ref{fig:graph_generation}(c) shows the state after the operation of this step. Then we add type constraints, time constraints and ordinal constraints in turn, and finally get the query graph as shown in Figure~\ref{fig:graph_generation}(d). 

Through the above procedure, we can obtain the candidate query graph set $G=\{g_1, g_2, ..., g_n\}$ for query graph selection.

\subsection{Query Graph Selection}
Due to the existence of ambiguity, the query graph generation may produce more than one, often hundreds of, candidate query graphs. Thus it is necessary to apply a matching operation to select the optimal query graph $g^*$ from the candidates. In this section, we first describe how to convert each $g$ in $G$ into sequence $g^s$. Then, we show how to encode the pair of $q$ and $g^s$. Finally, we describe the selection process. Inspired by the research in learning to rank~\citep{li2011learning,pirtoacua2019answering,han2020learning}, we select query graph with different ranking strategies, i.e., Pointwise Ranking, Pairwise Ranking, and Listwise Ranking.

\subsubsection{Transforming Query Graph into Sequence}
The process of converting query graph $g$ to sequence $g^s$ can be regarded as the disassembly process of constructing the query graph. When constructing the query graph, we first search the main path and then add four constraints of type, entity, time, and ordinal to the main path. Therefore, the whole query graph structure contains at most five components, which is much simpler compared with the general graph structure. And more importantly, each component has a fixed semantic meaning.

Considering the fixed structure of the query graph, we transform the query graph into the corresponding sequence according to the predefined sub-paths order. Specifically, we divide the query graph into different sub-paths according to different components. Through graph decomposition, we can get five sub-path sequences: TypePath, EntityPath, TimePath, OrdinalPath and MainPath. For example, the EntityPath corresponding to the entity constraint ``Prime minister" in Figure~\ref{fig:graph2sequence} is ``basic title prime minister.". Finally, the five sub-path sequences are combined to form the corresponding query graph sequence. It is worth noting that we use some additional tokens, [unused0-3], to separate different sub-path sequences. As shown in Figure~\ref{fig:graph2sequence}, the query graph sequence is ``people person. [unused0] basic title prime minister. [unused1] from after 1980. [unused2] height max 1. [unused3] spain governing officials -- office holder [A]'', in which `[A]' is the real answer string, not just unified padding.
\begin{figure*}
\centering
\includegraphics[width=17cm]{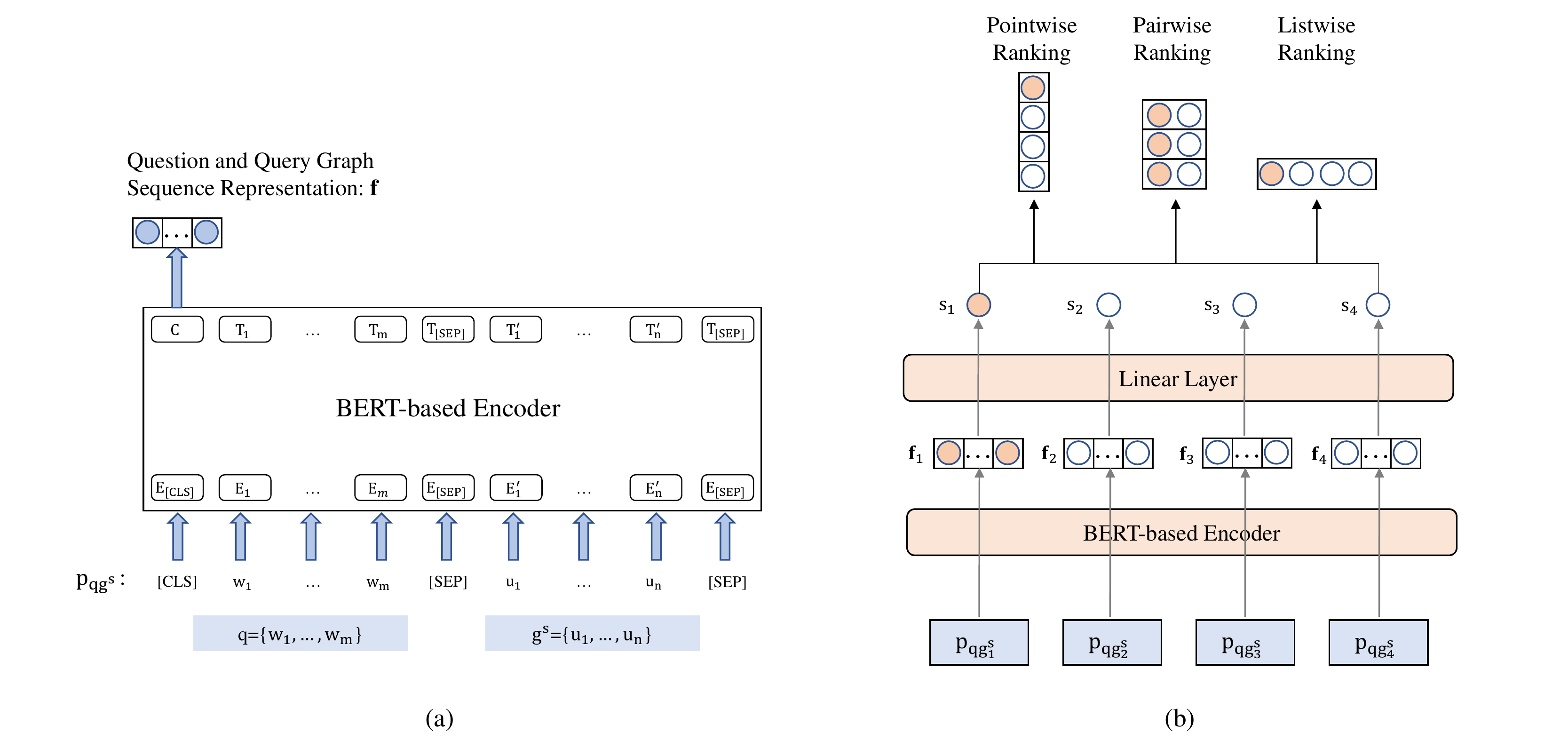}
\caption{(a) Query Graph and Question Encoding Framework. (b) Different Ranking Strategies Framework, where  ``$p_{qg^{s}_{1}}$" represents the sequence of the question and the positive query graph, ``$p_{qg^{s}_{2}}$", ``$p_{qg^{s}_{3}}$" and ``$p_{qg^{s}_{4}}$" represent three sequences of the question and negative query graphs.}
\label{fig:bert_frame}
\end{figure*}

\subsubsection{Encoding Query Graph Sequence and Question}

After query graph $g$ is converted into sequence $g^s$, the task of matching question $q$ and query graph $g$ becomes a task of matching question $q$ and query graph sequence $g^s$. This allows us to naturally use mature sequence encoding models, such as BERT~\citep{devlin2018bert}, GPT-3~\citep{brown2020language}, and so on, which can establish the interactive information of two sequences.

We choose the BERT architecture as our encoder, which has been widely used in natural language processing in recent years. BERT is a pre-trained language model based on the bidirectional Transformer architecture~\citep{vaswani2017attention}, which can be used to encode a single sentence or a sentence pair. To introduce the interactive information between the question and query graph sequence, we use the structure of encoding the sentence pair in BERT. The encoding framework is shown in Figure~\ref{fig:bert_frame}(a). Given the question $q=\{w_1,w_2,...,w_m\}$ and the query graph sequence $g^s=\{u_1,u_2,...,u_n\}$, we connect $q$ and $g^s$ through special tags to form the sentence pair, denoted as $p_{qg^s}=\{[CLS], w_1,...,w_m,[SEP],u_1,...,u_n,[SEP]\}$. Each candidate query graph $g$ in set $G$ can form a sentence pair $p_{qg^s}$ with the corresponding question $q$. Then the pairs are fed to BERT for encoding one by one. And we use the output of the $[CLS]$ node of BERT as the semantic representation of the question and query graph sequence, denoted as $\textbf{f}$. 

\subsubsection{Ranking Query Graphs}
In this section, we rank the candidates with three different strategies, namely Pointwise Ranking, Pairwise Ranking and Listwise Ranking, respectively. These three optimization methods correspond to three typical ranking strategies in information retrieval~\citep{li2011learning}. \citet{luo2018knowledge} use the pairwise strategy in their system and achieve a certain success. 

Before performing ranking, we preprocess the training data. According to whether the correct answer can be retrieved, the candidates can be grouped into two sets: $G^+$ and $G^-$, where $G^+$ includes the positive graphs and $G^-$ includes the negative ones. We use $g_i^+$ and $g_j^-$ to denote a positive graph and a negative graph, respectively. Each graph $g_i$ in the two sets is encoded as representation $\textbf{f}_i$. After that, $\textbf{f}_i$ is fed into the linear layer to get a score $s_i$ that indicates the possibility of $g_i$ being the optimal query graph.  
\paragraph{Pointwise Ranking.}
The Pointwise Ranking processes the graphs one by one. When ranking candidate query graphs, the query graphs to be sorted do not need have a perfect order, only the difference between positive and negative graphs. That is, we treat the query graph ordering problem as a simple binary classification task. As shown in Figure~\ref{fig:bert_frame}(b), the query graph $g_i$ in Pointwise Ranking is optimized independently.

For each candidate query graph $g_i$, it corresponds to the label $y_i\in\{1,0\}$, where `1' represents the label of a positive graph, and `0' represents the label of a negative graph. In the optimization process, we use the cross-entropy loss function to optimize and select the query graph with the highest score as the optimal query graph $g^*$. The optimized loss function is as follows,

\begin{equation}
    s^{'}_i = \frac{1}{1+e^{-s_i}},
    \label{con:sigmoid}
\end{equation}
\begin{equation}
    L_{point}=-\sum y_{i}log(s^{'}_i)+(1-y_i)log(1-s^{'}_i).
\end{equation}

\paragraph{Pairwise Ranking.}
The Pairwise Ranking can model the mutual connection between two candidate elements, and realize the ranking by continuously optimizing the relative rank of the two elements. When using the pairwise strategy to rank candidate query graphs, we regard the ranking problem as a problem of how to distinguish positive query graphs and negative query graphs. That is, we first construct the form of pairs of positive and negative graphs, and then the order of positive and negative graphs is optimized by using the pairwise strategy, as shown in Figure~\ref{fig:bert_frame}(b).

For each pair of positive and negative query graphs $(g_i^+,g_j^-)$, we can get the scores $s_i$ and $s_j$ through BERT and linear layer encoding, respectively. Then $s_i$ and $s_j$ are normalized to $s^{'}_i$ and $s^{'}_j$ by Equation (\ref{con:sigmoid}). We use the hinge loss optimization function to optimize the scores so that the difference between the scores of the positive and negative query graphs is maintained at a fixed value $\lambda$. The hinge loss is defined as follows,

\begin{equation}
L_{pair}=max\{0, \lambda - s^{'}_i + s^{'}_j\},
\end{equation}
where $\lambda$ is set to 0.5.
\paragraph{Listwise Ranking.}
The Listwise Ranking can also model the interconnection between all candidates, and can directly optimize the order of the entire candidate set. In this strategy, we construct a list of positive and negative graphs for global optimization. In the query graph selection, we don't care too much about the ranking among positive graphs or the ranking among negative graphs. Our goal of global optimization is to successfully rank the positive graph in the first place.

The application of Listwise in query graph ranking is slightly different from the method used in traditional information retrieval. When constructing the training data, we select each positive graph and a fixed number of negative graphs to form a list $C=\{{g}_{0}^{+}, {g}_{1}^{-}, {g}_{2}^{-}, …, {g}_{m}^{-}\}$, whose label is $\{y_0,y_1,y_2,...,y_m\}$. The score of group $C$ after BERT encoding and linear layer mapping is recorded as $\{s_{0}, s_{1}, s_{2}, …, s_{m}\}$. During the training, we design the following optimization objective function,
\begin{equation}
s^{'}_i=\frac{exp(s_{i})}{\sum_{i=0}^{m}exp(s_{i})},
\end{equation}
\begin{equation}
L_{list}=-\sum_{i=0}^{m}y_{i}log(s^{'}_i)+(1-y_i)log(1-s^{'}_i).
\end{equation}

\section{Experiments}

\subsection{Experimental Setup}
\paragraph{Datasets.}

\begin{table}
\centering
\begin{tabular}{lccc}
\hline \textbf{Dataset} & \textbf{train} & \textbf{validation} & \textbf{test} \\ \hline
WebQ & 3,023 & 755 & 2,032 \\
CompQ & 1,000 & 300 & 800 \\
\hline
\end{tabular}
\caption{\label{datasets split} The partitions of WebQuestions and ComplexQuestions. }
\end{table}

We conduct experiments on two widely-used datasets: WebQuestions (WebQ)~\citep{berant2013semantic}~\footnote{https://nlp.stanford.edu/software/sempre/} and ComplexQuestions (CompQ)~\citep{bao2016constraint}~\footnote{https://github.com/JunweiBao/MulCQA/
tree/ComplexQuestions}. The WebQ dataset
contains both simple questions (84\%) and complex reasoning questions (16\%), which is close to natural language questions used by people in daily life. The dataset contains 5,810 question-answer pairs. The CompQ dataset
is designed for complex question answering. The dataset contains a total of 2,100 question-answer pairs. Both WebQ and CompQ are divided into train set, validation set and test set, as shown in table~\ref{datasets split}. Both datasets use Freebase as the knowledge base~\footnote{https: //developers.google.com/freebase/}, which has been widely used in the KBQA systems.


\paragraph{Implementation Details.}
For encoding questions and query graphs, we utilize the BERT-base model. We choose the hyper-parameter settings according to the performance on the validation sets. Regarding the hyperparameters in the BERT-base model, we set the dropout ratio to 0.1 and the hidden size to 768. We use Adam as the optimizer and the learning rate is set to $5\times10^{-5}$. The maximum number of training epoch is set to 5. At the end of each epoch, we use the validation set to evaluate the model, and the model with the best performance on the validation set is selected as the final testing model. For performance evaluation, we report the average F1 score, as do in~\citet{berant2013semantic}.

One question left is how to construct training data for query graph selection. 
Given a question and its corresponding query graph candidates, we use the query graph whose answer has an F1 value greater than 0.1 as a positive graph and randomly sample negative graphs from the rest query graph candidates.




\subsection{Main Results}
 \begin{table}
\centering
 \begin{tabular}{lcc}
\hline \textbf{Method} & \textbf{WebQ (F1\%)} & \textbf{CompQ (F1\%)} \\ \hline
Pointwise & 52.4 & 38.4 \\
Pairwise & 53.7 & 42.7 \\
Listwise & 55.3 & 44.4\\
\hline
\end{tabular}
\caption{\label{rank-approach} The comparison results of the three ranking strategies on the test sets. }
\end{table}

\begin{table*}
\centering
\small \begin{tabular}{llcc}
\hline \textbf{Category} & \textbf{Method} & \textbf{WebQ(F1\%)} & \textbf{CompQ(F1\%)} \\ \hline
    & \citet{yih2015semantic} & 52.5 & - \cr
    &\citet{bao2016constraint} & 52.4 & 40.9 \cr
Using Query Graph& \citet{hu2018state} & 53.6 & - \cr
    & \citet{luo2018knowledge} & 52.7 & 42.8 \cr
    & \citet{lan2020query} & - & 43.3 \cr
\hline
    &\citet{berant2013semantic} & 36.4 & - \cr
Others    &\citet{jain2016question} & 55.6 & - \cr
    &\citet{chen2019bidirectional} & 51.8 & - \cr
    &\citet{xu2019enhancing} & 54.6 & - \cr
\hline

Our & Listwise & 55.3 & 44.4 \\
\hline
\end{tabular}
\caption{\label{previous-research-compare-table} The comparison results with previous works on the test sets of WebQuestions and ComplexQuestions. }
\end{table*}

Table~\ref{rank-approach} shows the comparison results of the three ranking strategies. From the table, we can see Listwise Ranking and Pairwise Ranking outperform the Pointwise Ranking. This fact indicates the necessity of modeling the inter-relations between query graph candidates. In addition, we also find that the superiority of Listwise Ranking and Pairwise Ranking are more significant on CompQ than WebQ, which is in line with our intuition that complex questions may require more information to disambiguate query graph candidates. Listwise Ranking yields the best result on both two datasets. The reason may be that Listwise considers more than two graphs at once which has a global view when optimization, compared with Pairwise.

Table~\ref{previous-research-compare-table} shows the comparison results between our system with Listwise Ranking and the previous works on the test sets of WebQ and CompQ, where category ``Using Query Graph'' includes the previous systems that use query graph and ``Others'' includes the previous ones that do not use query graph. From the table, we can see that our system yields the best result on CompQ and the second best on WebQ among all the systems. Especially, when compared with the approaches using query graph, our system achieves the best performance.

\subsection{Discussion and Analysis}
\subsubsection{Effect of Different Components in Query Qraph Selection}
\begin{table}
\centering
\begin{tabular}{lcc}
\hline \textbf{Sequence Info} & \textbf{WebQ (F1\%)} & \textbf{CompQ (F1\%)} \\ \hline
All Path & 55.3 & 44.4\\
w/o constrains & 53.7 & 42.3 \\
w/o answer & 54.3 & 43.6 \\
\hline
\end{tabular}
\caption{\label{ablation-table} The effect of different components in query graph sequence on query graph selection. }
\end{table}
When representing query graphs, we propose to transform a query graph into a sequence which is composed of sub-paths of different types. In order to explore the effect of different components on the final performance, we conduct experiments by removing some components from our Listwise system. The experimental results are presented in Table~\ref{ablation-table}, where ``All Path'' refers to our final system that includes the main path plus four constraint paths, ``w/o constrains'' refers to the system removing four constraint paths, and ``w/o answer'' refers to the system removing the answer string. The results show that by continuously accumulating different components, the system performance can be steadily improved. This indicates that all the components of the query graph sequence are useful for our system.

\subsubsection{Error Analysis}
In this section, we want to know the reasons why our system gives the wrong answers for many cases by error analysis.
If the candidate set generated by the query graph generation does not include the correct answer, our system is not possible to find it. Here we check the cases where the candidate set includes the correct answer but our system (Listwise) fails to find it. We randomly select 100 cases and check them manually. The errors are summarized as follows:

\textbf{Incorrect Query Graph Generation.}
There are some query graphs which can retrieve the correct answer, but actually are not correctly parsed. For example, the question ``where was david berkowitz arrested?'', the query graph generation provides ``david berkowitz places lived - location brooklyn, new york city'' as a candidate. The candidate graph can retrieve the correct answer, but in fact is not correct to the question. Of 100 cases, we find that this type of errors contains 45 cases. As for this type, we have to improve the performance of query graph generation and the coverage of KB. 

\textbf{Incorrect Query Graph Selection.}
As for the other cases, the candidate set includes the correct query graph which can perfectly retrieve the answer. However, our system still fails to find it. These errors can be grouped into two categories. The first one (40\%) is to select the graph that includes the incorrect relationship (main path) between the topic word and answer. The second one (15\%) is to select incorrect constraints.
To solve this type of errors, we may perform a deeper analysis to provide additional information for query graph selection. 

\subsubsection{Effect of Negative Examples on Ranking}
\begin{figure}
    \centering
    \includegraphics[width=8cm]{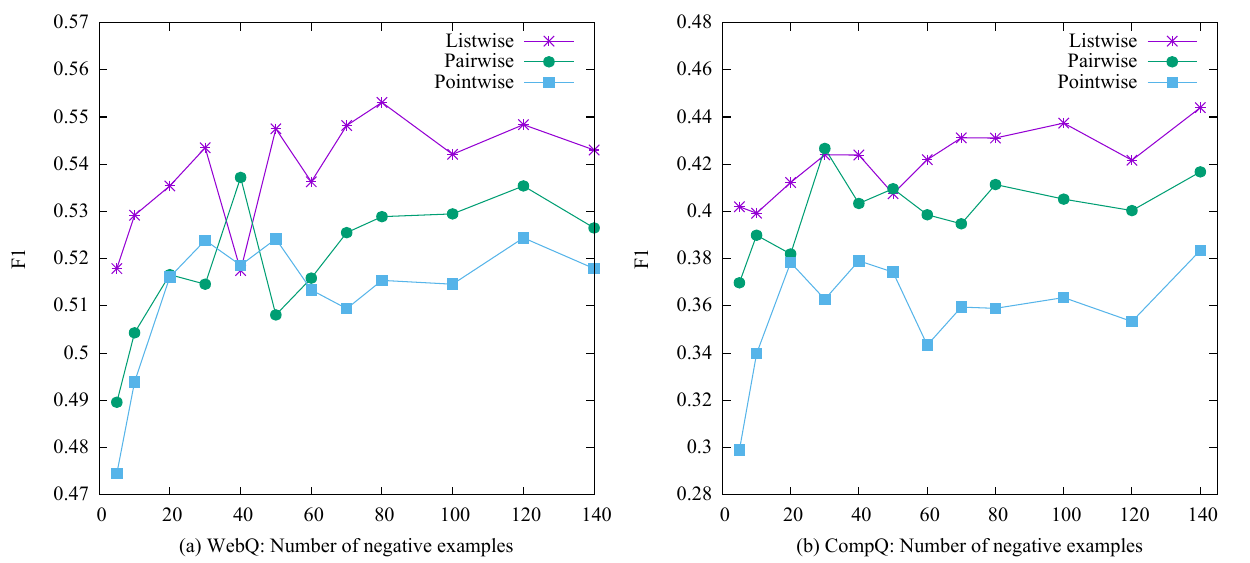}
    \caption{The performance of three ranking strategies under different numbers of negative examples.}
    \label{fig:compare_ranking}
\end{figure}

To further explore the characteristics of the three ranking strategies of Pointwise Ranking, Pairwise Ranking, and Listwise Ranking, we select different numbers of negative graphs to build the systems. The performance is shown in Figure~\ref{fig:compare_ranking}. From the figures, we find that when the number of negative graphs increases, the performance of all three systems first increases and then is in a relatively stable state. We also find that Listwise Ranking can yield a good performance with few negative samples. These facts indicate that we do not need too many negative graphs when training our systems.

\subsubsection{Case Study}
\begin{table}
\centering
\begin{tabular}{p{7cm}}
\hline
\textbf{Question1:} what type of breast cancer did sheryl crow have ?\\
\hline
\textbf{True:} sheryl crow condition meningioma.\\
\textbf{False:} sheryl crow films -- film breast cancer: the path of wellness \& healing. \\
\hline
\textbf{Question2:} what role did paul mccartney play in the beatles ?\\
\hline
\textbf{True:} member paul mccartney. the beatles member -- role backing vocalist, lead vocalist, bass. \\
\textbf{False:} the beatles (tv series) regular cast -- actor george harrison, john lennon, lance percival, paul frees, paul mccartney, ringo starr.\\
\hline
\end{tabular}
\caption{\label{case-table} The case study. }
\end{table}
We analyze some specific examples on which Listwise performs better than Pointwise. Two typical examples are listed in table~\ref{case-table}. For the question ``what type of breast cancer did sheryl crow have?'', the true answer should be a type of cancer. Listwise Ranking can determine that ‘condition’ is the correct relation, but Pointwise Ranking chooses the wrong path that contains ``breast cancer''. We argue that Listwise Ranking can better model the overall semantics of the sequence because it considers the relationship between the candidate query graphs during optimization, while Pointwise Ranking tends to focus on the semantics of local words. Besides, for the example ``what role did paul mccartney play in the beatles?'', the correct query graph contains the true entity constraint. But Pointwise Ranking chooses a path without the entity constraint. This indicates to some extent that Pointwise Ranking is not effective enough in identifying constrain paths.

\section{Related Work}
Information retrieval (IR) and semantic parsing (SP) based approaches are two mainstreams for knowledge base question answering. Among them, IR-based methods~\citep{yu2017improved,gupta2018retrieve,chen2019bidirectional,petrochuk2018simplequestions,zhao2019simple,saxena2020improving} obtain relevant candidate answers according to the topic entity and then rank the answers to obtain the final result. The core of IR-based approaches is to identify the KB relation paths that the question refers to~\citep{wu2019learning}. For example, ~\citet{dong2015question} use multi-column Convolutional Neural Networks (CNN) to encode questions and paths to the same vector space and calculate the similarity.~\citet{hao2017end} use Long Short-Term Memory (LSTM) instead of CNN for the same purpose.

Different from IR-based methods, SP-based approaches put more attention to the semantic analysis of the question~\citep{bao2016constraint}. The basic process of SP-based approaches is to parse the semantics of the question into some meaning representation and then map the meaning representation with KB~\citep{hu2017answering}. 
For example,~\citet{berant2013semantic} parse the question into $\lambda-DCS$, and then map it to the knowledge base through alignment and bridging operations to obtain answers.~\citet{sun2020sparqa} design a novel skeleton grammar to express complex questions and improve the ability to parse complex questions. Query graph is also a widely-used meaning representation in SP-based systems.~\citet{yih2015semantic} are the pioneer into query graph research for KBQA, which propose a staged query graph generation method for this task. Following this line,~\citet{luo2018knowledge} propose a complex query graph matching approach that simultaneously encodes multiple sub-paths to achieve better query graph representation. More recently,~\citet{lan2020query} propose a  method to expand multiple relations so that it can handle more complex questions. In contrast to previous works that mostly put focus on the representation of query graphs, we instead put focus on the phase of selecting the optimal query graph. 

\section{Conclusions}


We present a novel semantic matching approach based on semantic parsing to improve the performance of Knowledge Base Question Answering (KBQA). In this paper, the process of KBQA is divided into two steps: query graph generation and query graph selection, and we put focus on the second step. In our approach, we linearize the query graphs into sequences. Then, we use BERT to encode the pair of the query graph sequence and the question to obtain the semantic representation.  In addition, we select the optimal query graph with different ranking strategies, which take the relationship between candidate query graphs into consideration.  Experimental results on two benchmark datasets demonstrate the effectiveness of our proposed approach. Specifically, our best-performance system achieves the top performance on ComplexQuestions and the second best performance on WebQuestions.

\bibliography{aaai22}

\begin{thebibliography}{30}
\providecommand{\natexlab}[1]{#1}

\bibitem[{Auer et~al.(2007)Auer, Bizer, Kobilarov, Lehmann, Cyganiak, and
  Ives}]{auer2007dbpedia}
Auer, S.; Bizer, C.; Kobilarov, G.; Lehmann, J.; Cyganiak, R.; and Ives, Z.
  2007.
\newblock {DB}pedia: A nucleus for a web of open data.
\newblock In \emph{Proceedings of ISWC}, 722--735.

\bibitem[{Bao et~al.(2016)Bao, Duan, Yan, Zhou, and Zhao}]{bao2016constraint}
Bao, J.; Duan, N.; Yan, Z.; Zhou, M.; and Zhao, T. 2016.
\newblock Constraint-based question answering with knowledge graph.
\newblock In \emph{Proceedings of COLING}, 2503--2514.

\bibitem[{Berant et~al.(2013)Berant, Chou, Frostig, and
  Liang}]{berant2013semantic}
Berant, J.; Chou, A.; Frostig, R.; and Liang, P. 2013.
\newblock Semantic parsing on {F}reebase from question-answer pairs.
\newblock In \emph{Proceedings of EMNLP}, 1533--1544.

\bibitem[{Bollacker et~al.(2008)Bollacker, Evans, Paritosh, Sturge, and
  Taylor}]{bollacker2008freebase}
Bollacker, K.; Evans, C.; Paritosh, P.; Sturge, T.; and Taylor, J. 2008.
\newblock Freebase: {A} collaboratively created graph database for structuring
  human knowledge.
\newblock In \emph{Proceedings of SIGMOD}, 1247--1250.

\bibitem[{Brown et~al.(2020)Brown, Mann, Ryder, Subbiah, Kaplan, Dhariwal,
  Neelakantan, Shyam, Sastry, Askell, Agarwal, Herbert-Voss, Krueger, Henighan,
  Child, Ramesh, Ziegler, Wu, Winter, Hesse, Chen, Sigler, Litwin, Gray, Chess,
  Clark, Berner, McCandlish, Radford, Sutskever, and
  Amodei}]{brown2020language}
Brown, T.~B.; Mann, B.; Ryder, N.; Subbiah, M.; Kaplan, J.; Dhariwal, P.;
  Neelakantan, A.; Shyam, P.; Sastry, G.; Askell, A.; Agarwal, S.;
  Herbert-Voss, A.; Krueger, G.; Henighan, T.; Child, R.; Ramesh, A.; Ziegler,
  D.~M.; Wu, J.; Winter, C.; Hesse, C.; Chen, M.; Sigler, E.; Litwin, M.; Gray,
  S.; Chess, B.; Clark, J.; Berner, C.; McCandlish, S.; Radford, A.; Sutskever,
  I.; and Amodei, D. 2020.
\newblock Language Models are Few-Shot Learners.
\newblock \emph{arXiv:2005.14165}.

\bibitem[{Chen, Wu, and Zaki(2019)}]{chen2019bidirectional}
Chen, Y.; Wu, L.; and Zaki, M.~J. 2019.
\newblock Bidirectional attentive memory networks for question answering over
  knowledge bases.
\newblock In \emph{Proceedings of NAACL-HLT}, 2913--2923.

\bibitem[{Devlin et~al.(2019)Devlin, Chang, Lee, and
  Toutanova}]{devlin2018bert}
Devlin, J.; Chang, M.-W.; Lee, K.; and Toutanova, K. 2019.
\newblock {BERT}: Pre-training of deep bidirectional transformers for language
  understanding.
\newblock In \emph{Proceedings of NAACL-HLT}, 4171--4186.

\bibitem[{Dong et~al.(2015)Dong, Wei, Zhou, and Xu}]{dong2015question}
Dong, L.; Wei, F.; Zhou, M.; and Xu, K. 2015.
\newblock Question answering over {F}reebase with multi-column convolutional
  neural networks.
\newblock In \emph{Proceedings of ACL}, 260--269.

\bibitem[{Gupta, Chinnakotla, and Shrivastava(2018)}]{gupta2018retrieve}
Gupta, V.; Chinnakotla, M.; and Shrivastava, M. 2018.
\newblock Retrieve and re-rank: A simple and effective {IR} approach to simple
  question answering over knowledge graphs.
\newblock In \emph{Proceedings of FEVER}, 22--27.

\bibitem[{Han et~al.(2020)Han, Wang, Bendersky, and Najork}]{han2020learning}
Han, S.; Wang, X.; Bendersky, M.; and Najork, M. 2020.
\newblock Learning-to-Rank with {BERT} in {TF}-Ranking.
\newblock \emph{arXiv:2004.08476}.

\bibitem[{Hao et~al.(2017)Hao, Zhang, Liu, He, Liu, Wu, and Zhao}]{hao2017end}
Hao, Y.; Zhang, Y.; Liu, K.; He, S.; Liu, Z.; Wu, H.; and Zhao, J. 2017.
\newblock An end-to-end model for question answering over knowledge base with
  cross-attention combining global knowledge.
\newblock In \emph{Proceedings of ACL}, 221--231.

\bibitem[{Hu et~al.(2017)Hu, Zou, Yu, Wang, and Zhao}]{hu2017answering}
Hu, S.; Zou, L.; Yu, J.~X.; Wang, H.; and Zhao, D. 2017.
\newblock Answering natural language questions by subgraph matching over
  knowledge graphs.
\newblock \emph{IEEE Transactions on Knowledge and Data Engineering}, 30(5):
  824--837.

\bibitem[{Hu, Zou, and Zhang(2018)}]{hu2018state}
Hu, S.; Zou, L.; and Zhang, X. 2018.
\newblock A state-transition framework to answer complex questions over
  knowledge base.
\newblock In \emph{Proceedings of EMNLP}, 2098--2108.

\bibitem[{Jain(2016)}]{jain2016question}
Jain, S. 2016.
\newblock Question answering over knowledge base using factual memory networks.
\newblock In \emph{Proceedings of the NAACL Student Research Workshop},
  109--115.

\bibitem[{Kwiatkowski et~al.(2013)Kwiatkowski, Choi, Artzi, and
  Zettlemoyer}]{kwiatkowski2013scaling}
Kwiatkowski, T.; Choi, E.; Artzi, Y.; and Zettlemoyer, L. 2013.
\newblock Scaling semantic parsers with on-the-fly ontology matching.
\newblock In \emph{Proceedings of EMNLP}, 1545--1556.

\bibitem[{Lan and Jiang(2020)}]{lan2020query}
Lan, Y.; and Jiang, J. 2020.
\newblock Query graph generation for answering multi-hop complex questions from
  knowledge bases.
\newblock In \emph{Proceedings of ACL}, 969--974.

\bibitem[{Li(2011)}]{li2011learning}
Li, H. 2011.
\newblock Learning to rank for information retrieval and natural language
  processing.
\newblock \emph{Synthesis Lectures on Human Language Technologies}, 4(1):
  1--113.

\bibitem[{Liang(2013)}]{liang2013lambda}
Liang, P. 2013.
\newblock Lambda dependency-based compositional semantics.
\newblock \emph{arXiv:1309.4408}.

\bibitem[{Luo et~al.(2018)Luo, Lin, Luo, and Zhu}]{luo2018knowledge}
Luo, K.; Lin, F.; Luo, X.; and Zhu, K. 2018.
\newblock Knowledge base question answering via encoding of complex query
  graphs.
\newblock In \emph{Proceedings of EMNLP}, 2185--2194.

\bibitem[{Petrochuk and Zettlemoyer(2018)}]{petrochuk2018simplequestions}
Petrochuk, M.; and Zettlemoyer, L. 2018.
\newblock Simple{Q}uestions nearly solved: A new upperbound and baseline
  approach.
\newblock In \emph{Proceedings of EMNLP}, 554--558.

\bibitem[{P{\^\i}rtoac{\u{a}}, Rebedea, and
  Ruseti(2019)}]{pirtoacua2019answering}
P{\^\i}rtoac{\u{a}}, G.-S.; Rebedea, T.; and Ruseti, S. 2019.
\newblock Answering questions by learning to rank--Learning to rank by
  answering questions.
\newblock In \emph{Proceedings of EMNLP-IJCNLP}, 2531--2540.

\bibitem[{Saxena, Tripathi, and Talukdar(2020)}]{saxena2020improving}
Saxena, A.; Tripathi, A.; and Talukdar, P. 2020.
\newblock Improving multi-hop question answering over knowledge graphs using
  knowledge base embeddings.
\newblock In \emph{Proceedings of ACL}, 4498--4507.

\bibitem[{Sun et~al.(2020)Sun, Zhang, Cheng, and Qu}]{sun2020sparqa}
Sun, Y.; Zhang, L.; Cheng, G.; and Qu, Y. 2020.
\newblock SPARQA: Skeleton-Based Semantic Parsing for Complex Questions over
  Knowledge Bases.
\newblock In \emph{Proceedings of AAAI}, 8952--8959.

\bibitem[{Vaswani et~al.(2017)Vaswani, Shazeer, Parmar, Uszkoreit, Jones,
  Gomez, Kaiser, and Polosukhin}]{vaswani2017attention}
Vaswani, A.; Shazeer, N.; Parmar, N.; Uszkoreit, J.; Jones, L.; Gomez, A.~N.;
  Kaiser, {\L}.; and Polosukhin, I. 2017.
\newblock Attention is all you need.
\newblock In \emph{Proceeddings of NeurIPS}, 5998--6008.

\bibitem[{Wu et~al.(2019)Wu, Huang, Weng, Zheng, Zhang, Yan, and
  Chen}]{wu2019learning}
Wu, P.; Huang, S.; Weng, R.; Zheng, Z.; Zhang, J.; Yan, X.; and Chen, J. 2019.
\newblock Learning representation mapping for relation detection in knowledge
  base question answering.
\newblock In \emph{Proceedings of ACL}, 6130--6139.

\bibitem[{Xu et~al.(2019)Xu, Lai, Feng, and Wang}]{xu2019enhancing}
Xu, K.; Lai, Y.; Feng, Y.; and Wang, Z. 2019.
\newblock Enhancing key-value memory neural networks for knowledge based
  question answering.
\newblock In \emph{Proceedings of NAACL}, 2937--2947.

\bibitem[{Yang and Chang(2015)}]{yang2016s}
Yang, Y.; and Chang, M. 2015.
\newblock S-{MART} Novel tree-based structured learning algorithms applied to
  tweet entity linking.
\newblock In \emph{Proceedings of ACL}, 504--513.

\bibitem[{Yih et~al.(2015)Yih, Chang, He, and Gao}]{yih2015semantic}
Yih, S. W.-t.; Chang, M.-W.; He, X.; and Gao, J. 2015.
\newblock Semantic parsing via staged query graph generation: Question
  answering with knowledge base.
\newblock In \emph{Proceedings of ACL}, 1321--1331.

\bibitem[{Yu et~al.(2017)Yu, Yin, Hasan, Santos, Xiang, and
  Zhou}]{yu2017improved}
Yu, M.; Yin, W.; Hasan, K.~S.; Santos, C.~d.; Xiang, B.; and Zhou, B. 2017.
\newblock Improved neural relation detection for knowledge base question
  answering.
\newblock In \emph{Proceedings of ACL}, 571--581.

\bibitem[{Zhao et~al.(2019)Zhao, Chung, Goyal, and Metallinou}]{zhao2019simple}
Zhao, W.; Chung, T.; Goyal, A.; and Metallinou, A. 2019.
\newblock Simple Question Answering with Subgraph Ranking and Joint-Scoring.
\newblock In \emph{Proceedings of NAACL-HLT}, 324--334.

\end{thebibliography}
\end{document}